%% file: main.tex
\documentclass[sigconf,preprint,authorsversion]{acmart}

\renewcommand\footnotetextcopyrightpermission[1]{}
\setcopyright{none}

\AtBeginDocument{%
  }

\usepackage{cleveref}
\usepackage{algorithm}
\usepackage[noend]{algpseudocode}
\usepackage{mathtools}
\usepackage{multirow}
\usepackage{makecell}
\usepackage{subcaption}
\usepackage{microtype}

\input{notations}

\begin{document}

\title[Dominated Novelty Search: Rethinking Local Competition in Quality-Diversity]{Dominated Novelty Search:\texorpdfstring{\linebreak}{} Rethinking Local Competition in Quality-Diversity}


\author{Ryan Bahlous-Boldi}
\authornote{Both authors contributed equally to this research.}
\email{rbahlousbold@umass.edu}
\orcid{0000-0002-5789-0492}
\affiliation{%
    \institution{University of Massachusetts Amherst}
    \city{Amherst, MA}
    \country{USA}
}

\author{Maxence Faldor}
\authornotemark[1]
\email{m.faldor22@imperial.ac.uk}
\orcid{0000-0003-4743-9494}
\affiliation{%
    \institution{Imperial College London}
    \city{London}
    \country{United Kingdom}
}

\author{Luca Grillotti}
\email{luca.grillotti16@imperial.ac.uk}
\orcid{0000-0003-4539-8211}
\affiliation{%
    \institution{Imperial College London}
    \city{London}
    \country{United Kingdom}
}

\author{Hannah Janmohamed}
\email{hannah.janmohamed21@imperial.ac.uk}
\orcid{0000-0001-7997-8455}
\affiliation{%
    \institution{Imperial College London}
    \city{London}
    \country{United Kingdom}
}

\author{Lisa Coiffard}
\email{lisa.coiffard21@imperial.ac.uk}
\orcid{0009-0009-0557-9318}
\affiliation{%
    \institution{Imperial College London}
    \city{London}
    \country{United Kingdom}
}

\author{Lee Spector}
\email{lspector@amherst.edu}
\orcid{0000-0001-5299-4797}
\affiliation{%
    \institution{Amherst College}
    \city{Amherst, MA}
    \country{USA}
}

\author{Antoine Cully}
\email{a.cully@imperial.ac.uk}
\orcid{0000-0002-3190-7073}
\affiliation{%
    \institution{Imperial College London}
    \city{London}
    \country{United Kingdom}
}

\renewcommand{\shortauthors}{Bahlous-Boldi et al.}

\begin{abstract}
Quality-Diversity is a family of evolutionary algorithms that generate diverse, high-performing solutions through local competition principles inspired by natural evolution.
While research has focused on improving specific aspects of Quality-Diversity algorithms, surprisingly little attention has been paid to investigating alternative formulations of local competition itself --- the core mechanism distinguishing Quality-Diversity from traditional evolutionary algorithms.
Most approaches implement local competition through explicit collection mechanisms like fixed grids or unstructured archives, imposing artificial constraints that require predefined bounds or hard-to-tune parameters.
We show that Quality-Diversity methods can be reformulated as Genetic Algorithms where local competition occurs through fitness transformations rather than explicit collection mechanisms.
Building on this insight, we introduce Dominated Novelty Search, a Quality-Diversity algorithm that implements local competition through dynamic fitness transformations, eliminating the need for predefined bounds or parameters.
Our experiments show that Dominated Novelty Search significantly outperforms existing approaches across standard Quality-Diversity benchmarks, while maintaining its advantage in challenging scenarios like high-dimensional and unsupervised spaces.
\end{abstract}

\ccsdesc[500]{Computing methodologies~Genetic algorithms}

\keywords{Genetic Algorithms, Quality-Diversity, Open-endedness}
\begin{teaserfigure}
  \includegraphics[width=\textwidth]{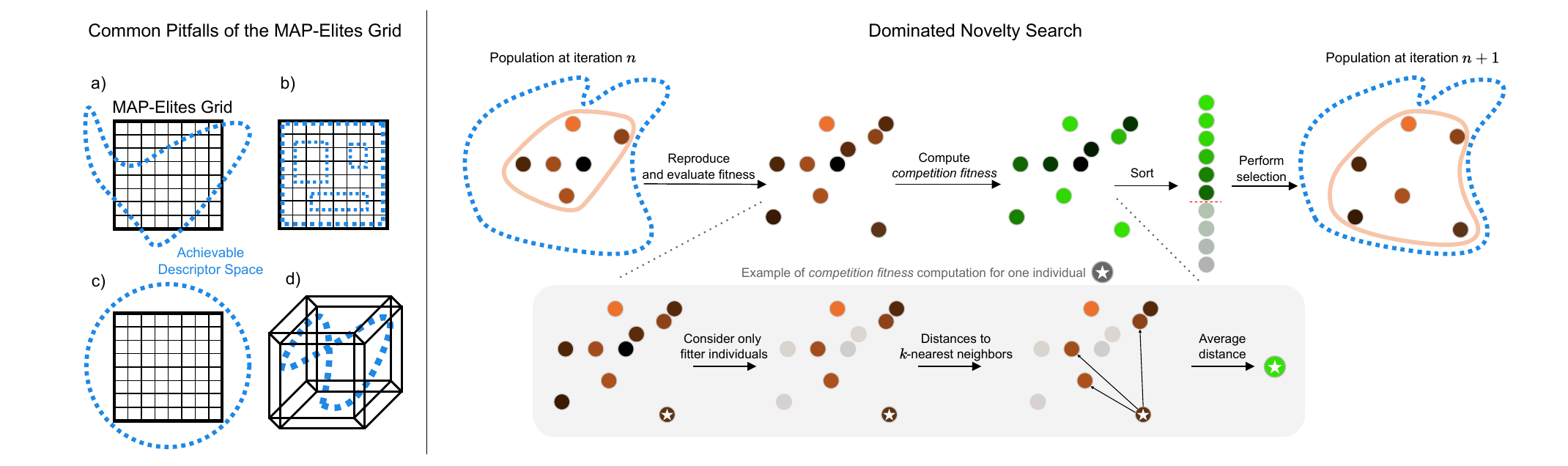}
  \caption{The MAP-Elites grid is limited by its ability to perform search when the achievable descriptor space a) has a complex topological shape, b) is discontinuous, c) is unbounded or d) is high-dimensional. In response, we propose \ours{}, a new local competition strategy for Quality-Diversity optimization.}
  \Description{}
  \label{fig:teaser}
\end{teaserfigure}

\received{29 January 2025}
\received[revised]{12 March 2009}
\received[accepted]{5 June 2009}

\maketitle

\pagestyle{plain}

\section{Introduction}
Evolution has created an extraordinary tapestry of life across Earth's biosphere. The human brain itself, with its 86 billion neurons forming trillions of synaptic connections, stands as a testament to the incredible complexity that can emerge from evolutionary processes.
Inspired by these biological principles, Genetic Algorithms~\citep{ga} (GA) harness the core mechanisms of evolution to solve complex optimization problems.
GAs iteratively evolve a population of solutions through a process of reproduction, evaluation and selection, over successive generations.
However, the selection implemented within GAs typically employs global competition within the population, often leading to convergence toward a single dominant solution that optimizes one specific objective.

This contrasts with natural evolution, where species have evolved to fill diverse environments, with competition occurring primarily between organisms within the same niche~\citep{darwin}.
This local competition allows multiple species to coexist and thrive simultaneously, each adapted to its specific environmental conditions, rather than converging to a single dominant species that would be globally optimal.
Inspired by this idea, Quality-Diversity~\citep{pugh_QualityDiversityNew_2016} (QD) is a family of evolutionary algorithms that harness local competition to evolve populations of diverse and high-performing solutions. In QD algorithms, each solution is characterized by both a fitness value measuring its performance and a descriptor vector capturing meaningful features about the solution. These descriptors embed solutions in a continuous space where distances between solutions can be computed, enabling competition to occur locally between similar solutions rather than globally across the entire population.

\begin{figure}
    \centering
    \includegraphics[width=\linewidth]{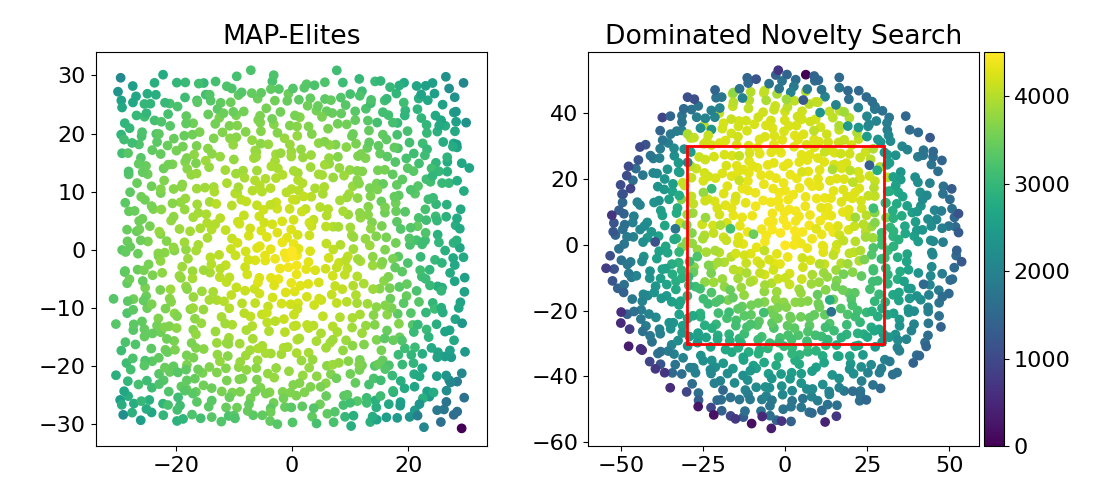}
    \caption{The descriptor specification problem. When trying to categorize the correct bounds for descriptors in QD optimization, one risks missing out on many quality-diverse solutions.}
    \Description{}
    \label{fig:descriptor-misspecification}
\end{figure}

MAP-Elites (ME) is a foundational QD algorithm that discretizes the descriptor space into a grid of cells~\citep{mouret_IlluminatingSearchSpaces_2015,vassiliades_UsingCentroidalVoronoi_2018}. Each cell maintains at most one solution, and competition occurs only between solutions mapped to the same cell. This grid-based approach provides a simple and effective way to structure local competition through spatial partitioning of the descriptor space.
However, this crude discretization comes with several fundamental limitations~\citep{cqd,vassiliades_ComparisonIlluminationAlgorithms_2017}.
First, grid-based approaches require defining the bounds of the descriptor space in advance, which is impractical in some cases. For instance, in unsupervised QD algorithms~\citep{cully_AutonomousSkillDiscovery_2019}, the descriptor space is learned during evolution, making it impossible to know its shape and bounds beforehand (\Cref{fig:teaser} a). Even with manually defined descriptors, the bounds of the descriptor space must often be determined experimentally, making the application of these algorithms arduous (\Cref{fig:teaser} c).
Second, even when bounds are known, some cells may map to regions of the descriptor space where no valid solutions can exist, as the reachable descriptor space rarely forms a perfect $n$-dimensional rectangular box (\Cref{fig:teaser} b). These permanently empty cells waste memory resources, ultimately reducing the effective size of the population.
Third, the MAP-Elites grid is known to struggle in high-dimensional spaces (\Cref{fig:teaser} d). Therefore, the commonly used grid shape rarely matches the natural shape of the solution space.

Quality-Diversity algorithms have explored approaches beyond fixed grid through unstructured archives~\citep{cully_AutonomousSkillDiscovery_2019,lehman_AbandoningObjectivesEvolution_2011,qd_framework}, which implement local competition enforcing a minimum distance threshold $l$ between individuals. While this creates dynamic niches that emerge naturally from the population structure rather than being predefined by fixed cells, it introduces its own heuristic through a fixed distance threshold that may not be appropriate across all regions of the descriptor space. This threshold parameter proves particularly challenging to tune, as its optimal value varies significantly across different domains and tasks, leading to the development of complex mitigation strategies such as container size control~\citep{csc}.

\begin{figure}
    \centering
    \includegraphics[width=1\linewidth]{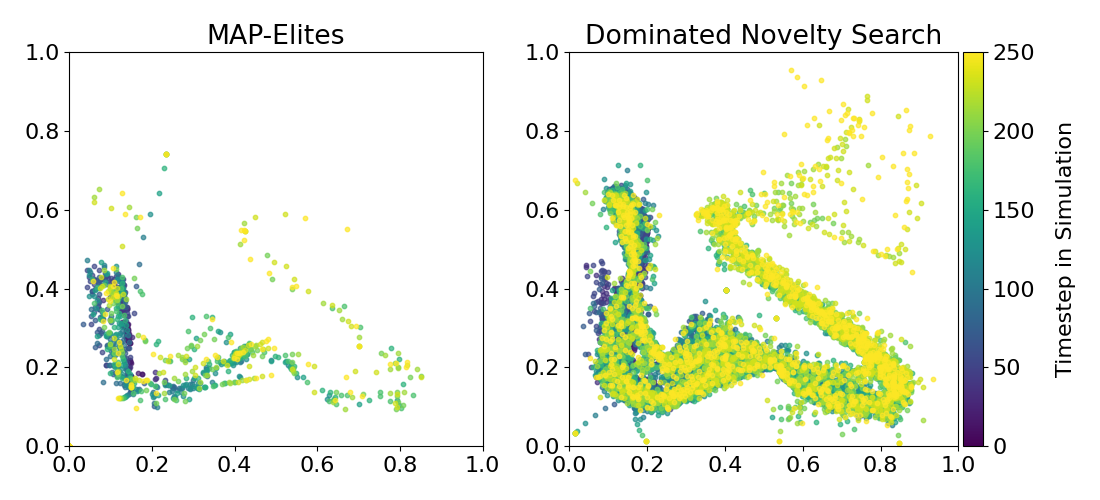}
    \caption{Scatter Plots of Kheperax ($n=30$), a high dimensional domain, where a solution's behavior descriptor is $30$ evenly spaced positions throughout the agent's trajectory. Due to this, a MAP-Elites grid would need to be 60 dimensions, with many dimension combinations being infeasible (e.g., being at the bottom left at the first step, and at the top right at the next)}
    \label{fig:high-dimensional-domain}
\end{figure}

The limitations of existing QD approaches point to a more fundamental question that remains largely unexplored in the Quality-Diversity literature: how can we design more sophisticated local competition strategies? While considerable research has focused on improving specific aspects of QD algorithms like reproduction~\citep{faldor_SynergizingQualityDiversityDescriptorConditioned_2024} or robustness~\citep{flageat2023UQD}, surprisingly little attention has been paid to investigating alternative formulations of local competition itself. The predominant grid-based competition in MAP-Elites and the distance-threshold approach in unstructured archives represent relatively simple competition rules, and the space of possible local competition strategies remains largely unexplored. This gap in our understanding is particularly significant given that local competition is the core mechanism distinguishing QD from traditional evolutionary algorithms. A deeper investigation of alternative competition strategies could unlock substantial improvements in QD algorithms' ability to generate diverse, high-performing solutions.

In this work, we introduce \ours{}, a Quality-Diversity algorithm that addresses these limitations through a novel local competition strategy. Our key insight is that Quality-Diversity algorithms can be viewed as Genetic Algorithms where local competition occurs through specialized fitness transformations, rather than through explicit collection mechanisms like grids or archives. This perspective differs from the traditional view of QD algorithms as distinct from GAs and enables us to derive new QD algorithms by modifying the local competition rules. \ours{} implements a novel competition strategy that dynamically adapts to the shape and structure of the descriptor space, eliminating the need for predefined bounds, grid structures, or fixed distance thresholds. Our approach can serve as a drop-in replacement for the grid mechanism in MAP-Elites, making it compatible with existing QD frameworks. Our experimental results demonstrate that \ours{} achieves state-of-the-art performance on standard QD benchmarks while introducing a new perspective on local competition in evolutionary algorithms.

\section{Related Work}
\label{sec:related-work}
\subsection{Local Competition in Genetic Algorithms}
Evolution's ability to generate diverse, well-adapted species has inspired numerous approaches in evolutionary computation to maintain population diversity. Early techniques focused on preventing premature convergence through mechanisms like fitness sharing~\citep{GeneticAlgorithmsSharing1987}, which enforces competition between similar solutions to encourage exploration of distant parts of the search space. \citet{alps} introduced the age-layered population structure where individuals compete only with others of similar genetic age, while \citet{hfc} introduced hierarchical fair competition to restrict competition to occur between solutions of similar fitness levels.

Several multi-objective and elitist algorithms promote diversity implicitly through their selection mechanisms as a natural consequence of search, rather than through an explicit diversity preservation mechanism. For example, NSGA-II~\citep{deb2002fast} maintains diversity through non-dominated sorting and crowding distance mechanisms, without needing a predefined behavior descriptor. Lexicase selection~\citep{helmuth_solving_2015}, which filters down individuals based on their performance on a large number of objectives that are considered in a random order, similarly maintains high levels of diversity~\citep{helmuth2016effects,boldi2024solving}.

A radical departure from these approaches came with Novelty Search~\citep{lehman_AbandoningObjectivesEvolution_2011}, which completely abandons the objective function and instead rewards solutions solely for being behaviorally different from their predecessors. By introducing the concept of behavioral distance as a fundamental measure for competition, Novelty Search laid crucial groundwork for new forms of local competition that would later be integrated with objective-based optimization in Quality-Diversity algorithms. 

\subsection{Quality-Diversity}
Quality-Diversity algorithms address the challenge of discovering diverse, high-performing solutions through local competition mechanisms~\citep{pugh_QualityDiversityNew_2016}. MAP-Elites~\citep{mouret_IlluminatingSearchSpaces_2015} formalizes this through a grid-based architecture where solutions compete only within their behavioral cell, maintaining the highest-performing individual in each niche. To enhance scalability in high-dimensional behavior spaces, Centroidal Voronoi Tessellations~\citep{vassiliades_UsingCentroidalVoronoi_2018} adaptively partition the space into well-distributed regions. While these approaches prove effective, particularly in robotics applications, they rely on predefined bounds and potentially arbitrary discretization of the behavior space.

Unstructured archives offer an alternative approach to local competition. Novelty Search with Local Competition~\citep{lehman_EvolvingDiversityVirtual_2011} and AURORA~\citep{cully_AutonomousSkillDiscovery_2019,qd_framework} make solutions compete only against their behavioral neighbors through a distance-threshold mechanism. While these methods eliminate the need for predefined cells, they introduce new challenges through fixed distance parameters that require careful tuning and often need complex adaptation mechanisms like container size control~\citep{csc}. Cluster-Elites~\citep{vassiliades_ComparisonIlluminationAlgorithms_2017} attempts to address these limitations by dynamically spreading centroids across the actual shape of the discovered solutions rather than within a predefined hyperrectangle, allowing it to better adapt to non-convex behavioral spaces.

Our method, \ours{}, introduces a novel approach to local competition that rewards solutions that either outperform their neighbors or find unique behaviors compared to better-performing solutions. Rather than using spatial partitioning or distance thresholds, \ours{} creates an emergent competition structure based on relative solution performance and proximity. This approach naturally adapts to the distribution of solutions without requiring predefined bounds or parameters, making it particularly effective for domains with unknown or complex behavioral spaces.

\section{Background}
\subsection{Genetic Algorithms}
\label{sec:background-ga}
Genetic Algorithms (GAs) implement artificial evolution by maintaining a population of candidate solutions and iteratively selecting the most promising ones~\citep{ga}. At each generation, GAs create new solutions through reproduction operators like crossover and mutation, evaluate their fitness, and select which solutions survive to the next generation. A common implementation of GAs is outlined in \Cref{alg:ga}, though many variants exist in the literature.

\begin{algorithm}
\caption{Genetic Algorithm}
\label{alg:ga}
\small
\begin{algorithmic}
\Require population size $N$, reproduction batch size $B$
\State Initialize population $\population$ with fitness $\fitnesses$
\For{each generation}
    \State $\population^\prime \gets \textsc{reproduction}(\population, \fitnesses)$\Comment{Generate $B$ offspring}
    \State $\mathrlap{\population}\hphantom{\population^\prime} \gets \textsc{concat}(\population, \population^\prime)$\Comment{Add offspring to population}
    \State $\mathrlap{\fitnesses}\hphantom{\population^\prime} \gets \textsc{evaluation}(X)$\Comment{Evaluate fitness}
    \State $\mathrlap{\population}\hphantom{\population^\prime} \gets \textsc{selection}(\population, \fitnesses)$\Comment{Keep top-$N$ individuals}
\EndFor
\end{algorithmic}
\end{algorithm}

The GA maintains a population of $N$ candidate solutions $\population = (\individual_i)_{i=1}^N \in \mathbb{R}^{N \times n}$ and their associated fitness scores $\fitnesses = (\fitness_i)_{i=1}^N \in \mathbb{R}^N$ across generations.
In each iteration, the algorithm generates $B$ new solutions $\population^\prime = \textsc{reproduction}(\population, \fitnesses)$ through genetic operators such as crossover and mutation.
These newly generated solutions are added into the existing population via $\textsc{concat}(\population, \population^\prime)$, resulting in an expanded set of $N+B$ solutions.
The generation concludes with the selection operator, which sorts solutions according to their competition fitness $\fitnesses$ and preserves the top-$N$ performers through truncation selection.

\subsection{Quality-Diversity}
\label{sec:background-qd}
Quality-Diversity algorithms extend traditional evolutionary approaches by simultaneously optimizing both the quality and diversity of solutions. Unlike standard GAs that focus solely on fitness optimization, QD algorithms introduce descriptors that characterize meaningful features of solutions. Each solution in QD is evaluated based on both its performance (fitness) and its characteristics (descriptor), represented by a descriptor vector $\descriptor \in \descriptorSpace \subset \mathbb{R}^D$.

QD algorithms evolve a population of $N$ solutions $\population = (\individual_i)_{i=1}^N \in \mathbb{R}^{N \times n}$, maintaining both fitness values $\fitnesses = (\fitness_i)_{i=1}^N \in \mathbb{R}^N$ and descriptors $\descriptors = (\descriptor_i)_{i=1}^N$ across generations. This integration of descriptors represents a fundamental shift in evolutionary optimization, as it actively seeks to discover multiple high-performing solutions that exhibit different characteristics. This diversity-aware optimization process has led to several algorithmic variants, each employing different mechanisms to organize and maintain their solution collections. These variants primarily differ in how they structure their populations~\citep{qd_framework} --- through grids, archives, or other organizational schemes --- which we detail in the following sections.

\subsubsection{Grid}
\label{sec:background-me}
MAP-Elites~\citep{mouret_IlluminatingSearchSpaces_2015} is a QD algorithm that performs local competition through discretization of the descriptor space into a grid structure.
Each cell in the grid is characterized by a centroid $\centroid_i$, forming $\centroids = (\centroid_i)_{i=1}^N$.
At most one solution can be stored in each cell and competition is restricted to solutions mapped to the same cell~\citep{mouret_IlluminatingSearchSpaces_2015,vassiliades_UsingCentroidalVoronoi_2018}.
This mechanism ensures selective pressure within each descriptor space partition while maintaining diversity across cells.
While grids now commonly use CVT for better scaling~\citep{vassiliades_UsingCentroidalVoronoi_2018}, we refer to both traditional grids and CVT-based approaches as grid-based methods.

\subsubsection{Unstructured Archive}
\label{sec:background-unstructured}
Quality-Diversity algorithms with unstructured archives~\citep{qd_framework,lehman_AbandoningObjectivesEvolution_2011} extend the original MAP-Elites algorithm, offering a flexible framework for handling unknown descriptor spaces~\citep{cully_AutonomousSkillDiscovery_2019,csc}.
These archives implement local competition through a minimum distance threshold $l$ between solutions in the descriptor space. For each solution, the algorithm calculates its novelty score as the mean distance to its $k$ nearest neighbors in the descriptor space. The selection process then considers both the fitness value and this novelty measure, comparing each solution against its closest neighbors to determine which solutions to retain.
This flexibility makes unstructured archives particularly valuable in scenarios where the bounds and topology of the descriptor space are unknown beforehand, such as in unsupervised Quality-Diversity optimization~\citep{cully_AutonomousSkillDiscovery_2019,csc,qd_framework}.
In the remainder of this paper, as multiple archives are unstructured, we will refer to this specific variant as Threshold-Elites (TE).

\subsubsection{Cluster-Elites}
\label{sec:background-cluster-elites}
Cluster-Elites~\citep{vassiliades_ComparisonIlluminationAlgorithms_2017} represents an innovative approach to Quality-Diversity optimization that addresses the limitations of fixed-grid methods while maintaining their structural advantages.
Unlike traditional MAP-Elites, which requires predefined bounds for the descriptor space, Cluster-Elites adaptively discovers and organizes the solution space through dynamic centroid allocation.
When using this algorithm, the population is divided into two distinct subpopulations: $\centroids$, which serves as centroids for the CVT grid, and $\population$, containing solutions that undergo local competition within the CVT grid.
A key advantage of this method lies in separating two core objectives: maximizing diversity through strategic centroid selection, while improving solution quality through local competition within grid cells.

\section{Method}
\label{sec:method}
Quality-Diversity optimization has traditionally relied on explicit mechanisms like archives and grids to maintain diversity. Here, we propose a fundamentally different approach that reformulates these mechanisms through the lens of fitness transformations.
In~\Cref{sec:qd-as-ga-lc}, we reformulate QD algorithms as GAs with local competition, demonstrating how fitness transformations can replace traditional archive or grid-based mechanisms. Subsequently, in~\Cref{sec:dns}, we present a specific instantiation of this framework through a novel fitness transformation that enables effective Quality-Diversity optimization. While we focus on one particular implementation, this framework opens possibilities for exploring alternative competition functions, including learned competition mechanisms --- though such extensions lie beyond the scope of this work.

\subsection{Quality-Diversity as a Genetic Algorithm with Local Competition}
\label{sec:qd-as-ga-lc}
Traditional Quality-Diversity algorithms typically maintain diversity through explicit collection mechanisms such as grids (ME) or archives (NSLC). We present a novel perspective on Quality-Diversity algorithms by reformulating them as GAs augmented with specialized fitness transformations that implement local competition. This perspective shifts focus from the storage mechanisms to the underlying competition dynamics that drive evolutionary search.

\begin{algorithm}
\caption{Quality-Diversity}
\label{alg:qd}
\small
\begin{algorithmic}
\Require population size $N$, reproduction batch size $B$
\State Initialize population $\population$ with fitness $\fitnesses$ and descriptors $\descriptors$
\For{each generation}
    \State $\mathrlap{\population^\prime}\hphantom{\fitnesses, \descriptors} \gets \textsc{reproduction}(\population, \fitnesses)$\Comment{Generate $B$ offspring}
    \State $\mathrlap{\population}\hphantom{\fitnesses, \descriptors} \gets \textsc{concat}(\population, \population^\prime)$\Comment{Add offspring to population}
    \State $\mathrlap{\fitnesses, \descriptors}\hphantom{\fitnesses, \descriptors} \gets \textsc{evaluation}(X)$\Comment{Evaluate fitness of population}
    \State {\color{green}$\mathrlap{\tilde{\fitnesses}}\hphantom{\fitnesses, \descriptors} \gets \textsc{competition}(\fitnesses, \descriptors)$\Comment{Local competition}}
    \State $\mathrlap{\population}\hphantom{\fitnesses, \descriptors} \gets \textsc{selection}(\population, \tilde{\fitnesses})$\Comment{Keep top-$N$ individuals}
\EndFor
\end{algorithmic}
\end{algorithm}

In this framework, Quality-Diversity algorithms closely mirror the structure of GAs, following the same core operations of reproduction, evaluation, and selection. The key distinction lies in how QD algorithms determine which solutions survive to the next generation. Rather than using raw fitness values directly, QD introduces a \emph{competition} function that transforms these values into \emph{competition fitness} scores: $\tilde{\fitnesses} = \textsc{competition}(\fitnesses, \descriptors)$.
This competition function considers both the raw fitness $\fitnesses$ of solutions and their relative positions in descriptor space $\descriptors$, creating competition fitness values $\tilde{\fitnesses}$ that drive selection.

The selection mechanism itself remains identical to traditional GAs --- solutions are ranked by their competition fitness $\tilde{\fitnesses}$ and truncated to maintain a fixed population size $N$. However, by incorporating descriptor-space proximity into the competition fitness, this approach maintains diverse solutions across different niches while driving optimization within each niche. Our reformulation reveals that the fundamental innovation of QD algorithms is not their collection mechanisms (grids, archives) but rather how they transform fitness based on local competition. As shown in \Cref{alg:qd}, the only difference from the standard GA (\Cref{alg:ga}) is the addition of the competition step that computes $\tilde{\fitnesses}$. This perspective unifies existing QD approaches and suggests that new algorithms can be derived by designing novel competition functions.

This reformulation reveals that existing QD algorithms can be understood as implementing different competition functions. For example, in \Cref{sec:me,sec:threshold} we show that MAP-Elites and Theshold-Elites can be cast within this framework. Our framework suggests that the space of possible local competition strategies extends far beyond existing approaches. By viewing QD algorithms through the lens of competition functions, we open possibilities for designing novel algorithms that can better capture the nuances of local competition in different domains. This perspective also provides a unified theoretical foundation for understanding and comparing different QD approaches, potentially leading to more effective algorithms for generating diverse, high-performing solutions.

\subsubsection{MAP-Elites}
\label{sec:me}
MAP-Elites implements competition through its grid structure, where solutions compete only within their assigned cells. For a population with fitness values $\fitnesses$ and descriptors $\descriptors$, the $\textsc{competition}$ function computes $\tilde{f}$ as follows:
\begin{enumerate}
    \item Assignment of individuals to their nearest centroid based on descriptor values, partitioning the population into a grid.
    \item Within each cell $i$, preservation of only the highest-fitness individual $k$ by maintaining its original fitness $\tilde{\fitness}_k = \fitness_k$, while eliminating others through $\tilde{\fitness}_j = -\infty$, for $j \neq k$.
\end{enumerate}

\subsubsection{Threshold-Elites}
\label{sec:threshold}
In Threshold-Elites, the $\textsc{competition}$ function computes $\tilde{f}$ as follows:
\begin{enumerate}
    \item For each solution $\individual_i$, find its nearest living neighbor $\individual_j$ (where $j<i$ and $\fitness_j>-\infty$).
    \item If $\norm{\descriptor_i - \descriptor_j} > \threshold$, then solution $i$ retains its original fitness value: $\tilde{\fitness}_i = \fitness_i$.
    \item If $\norm{\descriptor_i - \descriptor_j} \leq \threshold$, then novelty scores are computed for both solutions. When $\individual_i$ improves upon $\individual_j$ in terms of fitness and novelty, it replaces $\individual_j$ by setting $(\tilde{\fitness}_i, \tilde{\fitness}_j) = (\fitness_i, -\infty)$. Otherwise, $\individual_j$ is retained and $\tilde{\fitness}_i=-\infty$.
\end{enumerate}


\subsection{\ours{}}
\label{sec:dns}
Building on~\Cref{sec:qd-as-ga-lc}, we introduce \ours{}, a Quality-Diversity optimization method that implements a new competition function. Unlike MAP-Elites' rigid grid-based partitioning or Threshold-Elites' fixed distance threshold, our method dynamically adapts to the structure of the solution space through a competition mechanism based on relative distances between solutions of different fitness levels.

The key insight behind \ours{} is that solutions should be eliminated from the population if they are both inferior in fitness and behaviorally similar to existing solutions. We formalize this intuition through a competition function that computes modified fitness values $\tilde{\fitnesses}$ based on how distant solutions are from other better-performing solutions. Given a population with fitness values $\fitnesses = (\fitness_i)_{i=1}^N$ and descriptors $\descriptors = (\descriptor_i)_{i=1}^N$, the competition function operates in three steps:
\begin{enumerate}
    \item For each solution $i$, identify all solutions with superior fitness:
        $$\mathcal{D}_i = \{j \in \{1,\ldots,N\} \mid \fitness_j > \fitness_i\}$$
    
    \item Compute pairwise distances in descriptor space between the solution $i$ and all the fitter solutions:
        $$d_{ij} = \norm{\descriptor_i - \descriptor_j} \quad \forall j \in \mathcal{D}_i$$
    
    \item Calculate the competition fitness $\tilde{\fitness}_i$ as the \emph{dominated novelty score} --- the average distance to the $k$-nearest-fitter solutions:
        $$\tilde{\fitness}_i =
            \begin{cases}
            \frac{1}{k}\sum_{j \in \mathcal{K}_i} d_{ij} & \text{if } |\mathcal{D}_i| > 0\\
            +\infty & \text{otherwise}
            \end{cases}
        $$
        where $\mathcal{K}_i$ contains the indices of the $k$ solutions in $\mathcal{D}_i$ with smallest distances $d_{ij}$ to solution $i$. If $|\mathcal{D}_i| < k$, we use all available fitter solutions.
\end{enumerate}

This competition mechanism creates two complementary evolutionary pressures: individuals must either improve their fitness or discover distinct behaviors that differ from better-performing solutions. Solutions that have no fitter neighbors ($\mathcal{D}_i = \emptyset$) receive an infinite competition fitness, ensuring their preservation in the population. The parameter $k$ controls the locality of competition by determining how many nearby fitter solutions influence each solution's competition fitness. This competition function is used as part of Algorithm~\ref{alg:qd}, where individuals with above median scores are then selected for reproduction in a GA.

A key advantage of \ours{} is that it requires no prior knowledge of descriptor space bounds or structure, as it naturally adapts to the reachable regions of the descriptor space. This property makes our method particularly effective for scenarios where the bounds of valid solutions are unknown or when working with learned descriptors, as we demonstrate in our experimental results (Section~\ref{sec:experiments}).

\section{Experiments}
\label{sec:experiments}
To validate the effectiveness of \ours{}, we perform extensive experimental evaluations across multiple domains and compare against standard QD algorithms. Our experimental investigation addresses three fundamental questions:
\begin{itemize}
    \item How does \ours{} perform compared to established QD algorithms on standard continuous control tasks with well-defined descriptor spaces (\Cref{sec:experiments-continuous})?
    \item Can \ours{} effectively handle environments with discontinuous descriptor spaces where traditional grid-based approaches struggle (\Cref{sec:experiments-discontinuous})?
    \item How well does \ours{} scale to high-dimensional descriptor spaces where defining explicit bounds becomes impractical (\Cref{sec:experiments-high-dim})?
\end{itemize}
Each experiment is repeated 10 times with different seeds. We employ the Mann-Whitney $U$ test coupled with Holm-Bonferroni correction to establish statistical significance across multiple comparisons. Our experiments utilize the JAX framework~\citep{jax} for efficient parallel computation and the QDax library~\citep{qdax} for QD algorithms. All experiments were conducted on NVIDIA A100 GPUs. We have made our repository available \href{https://github.com/adaptive-intelligent-robotics/Dominated-Novelty-Search}{here}.

\subsection{Tasks}
To evaluate the broad applicability of \oursShort{}, we conducted experiments across a diverse set of tasks featuring distinct descriptor space characteristics. Our evaluation spans two main categories of tasks: continuous control locomotion tasks implemented in the Brax physics simulator~\citep{brax2021github}, and maze navigation tasks based on the deceptive maze environment originally introduced by \citet{lehman_AbandoningObjectivesEvolution_2011} and reimplemented in JAX by \citet{grillotti23Kheperax}. Table~\ref{tab:tasks} summarizes the key properties of each task.

\begin{table}[h]
\caption{Overview of tasks and environments.}
\label{tab:tasks}
\small
\centering
\newdimen\length
\length=1.2cm
\begin{tabular}{l@{\hspace{1mm}}|@{\hspace{1mm}}c@{\hspace{0.5mm}}c@{\hspace{0.5mm}}c@{\hspace{0.5mm}}c@{\hspace{0.5mm}}c}
    \toprule
    & \textsc{Walker} & \textsc{Ant} & \textsc{Ant Blocks} & \multicolumn{2}{c}{\textsc{Kheperax}}\\
    & \includegraphics[height=\length, width=\length]{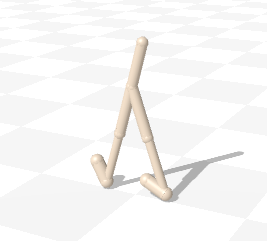} & 
     \includegraphics[height=\length, width=\length]{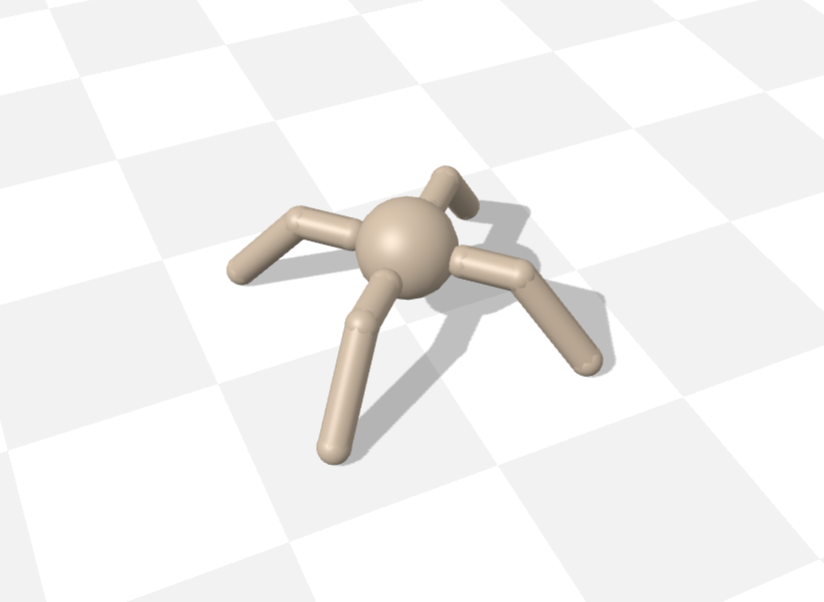} &
     \includegraphics[height=\length, width=\length]{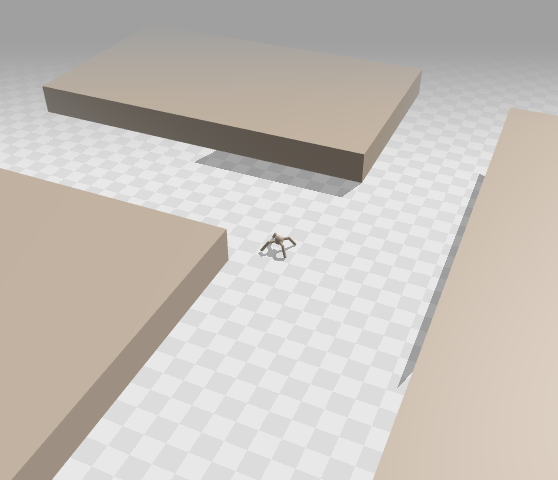} &
     \multicolumn{2}{c}{\includegraphics[height=\length, width=\length]{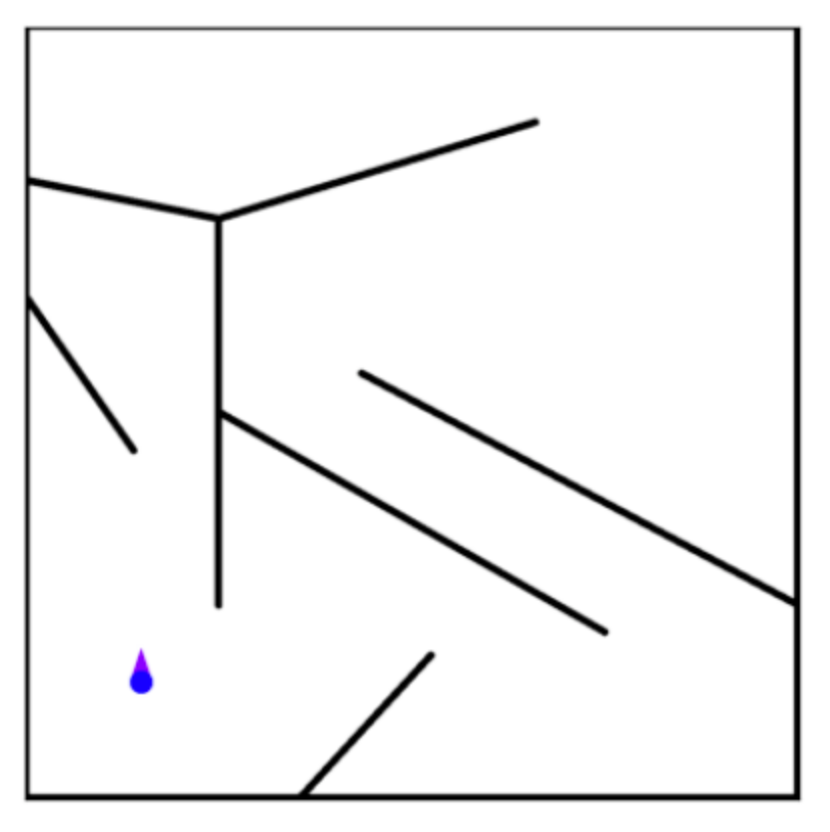}}\\
    \midrule
    \makecell[l]{\textsc{State}} & \multicolumn{3}{c}{\makecell{Joints velocity}} & \multicolumn{2}{c}{\makecell{Laser ranges}}\\
    \makecell[l]{\textsc{Action}} & \multicolumn{3}{c}{\makecell{Joints torque}} & \multicolumn{2}{c}{\makecell{Wheels velocity}}\\
    \makecell[l]{\textsc{Desc}} & \makecell{Feet Contact} & \multicolumn{2}{c}{\makecell{Final position}} & \makecell{$n$ points} & \makecell{Encoding}\\
    \makecell[l]{\textsc{Desc Space}} & \makecell{$[0, 1]^2$} & \makecell{$[-30, 30]^2$} & \makecell{$[-30, 30]^2$} & \makecell{$([0, 1]^2)^{n}$} & \makecell{$\mathbb{R}^{10}$}\\
    \bottomrule
\end{tabular}
\end{table}

\subsubsection{Walker and Ant}
\label{sec:experiments-continuous}
The continuous control locomotion tasks challenge agents to move efficiently. The agent's state space includes the position and orientation of its base and the angular positions and velocities of its joints. The action space consists of continuous torque commands that can be applied to each actuated joint.
In the Walker task, a bipedal agent must maximize its forward velocity while minimizing energy consumption. The descriptor space for this task captures the contact patterns between the agent's feet and the ground, measured as the contact rate for each foot.
In the Ant task, a quadrupedal agent must minimize energy consumption without any forward velocity reward. In this environment, the descriptor space is defined by the agent's final position, creating a continuous 2D descriptor space.

\subsubsection{Ant Blocks}
\label{sec:experiments-discontinuous}
The Ant Blocks environment extends the base Ant task by introducing static obstacles throughout the environment. As in the base task, the quadrupedal agent must optimize for energy-efficient movement while its final position defines the descriptor space. However, the presence of blocks creates natural barriers in the environment, resulting in a discontinuous descriptor space where certain regions become unreachable.

This discontinuity presents a particular challenge for traditional grid-based approaches like MAP-Elites, which predefine descriptor space bounds and allocate computational resources uniformly across this space. When grid cells are allocated to unreachable regions blocked by obstacles, the effective population size is reduced, potentially limiting the algorithm's ability to discover and maintain diverse solutions in the achievable regions of the descriptor space.

\subsubsection{Kheperax}
\label{sec:experiments-high-dim}
To evaluate performance in high-dimensional descriptor spaces, we employ the Kheperax environment~\citep{grillotti23Kheperax}, that requires agents to traverse complex mazes. Unlike traditional maze tasks that only consider the agent's final position, we characterize behavior using the agent's trajectory through the environment. Specifically, we construct a \emph{high-dimensional} descriptor by sampling the agent's position at regular intervals throughout its movement, creating a comprehensive representation of its navigation strategy.
This trajectory-based approach enables a richer analysis of agent behavior but introduces significant computational challenges. The resulting high-dimensional descriptor space makes traditional QD methods less effective due to the curse of dimensionality~\citep{cully_AutonomousSkillDiscovery_2019}. Grid-based approaches particularly struggle, as defining meaningful bounds becomes increasingly difficult with each additional dimension, and the number of cells grows exponentially with the descriptor dimensionality.

By selecting these tasks, we aim to demonstrate \oursShort{}'s capability to handle environments with continuous, discontinuous, and high-dimensional descriptor spaces, showcasing its robustness and adaptability across various scenarios.

\subsection{Evaluation Metrics}
\paragraph{Metrics} We employ two complementary metrics to evaluate algorithm performance: the Quality-Diversity score (QD score) and coverage. The QD score, a standard metric in QD research, measures the cumulative fitness across all solutions in the population, capturing both the quality of individual solutions and the breadth of behaviors discovered. Coverage quantifies purely the diversity aspect by measuring the proportion of descriptor space niches that contain viable solutions.

Traditional implementations of these metrics assume a structured archive, typically the grid used in MAP-Elites. However, this assumption creates challenges when evaluating algorithms with unstructured populations, where solutions are not explicitly organized into predefined niches. In such cases, direct computation of these metrics could emphasize solution quality at the expense of behavioral diversity, particularly for the QD score which might favor clusters of solutions in high-performing regions.

\paragraph{Projected metrics} To ensure comparability across different methods, especially those without inherent archive structure, we project the individuals onto a predefined grid-based archive, similar to that used by ME, to compute metrics. This approach is similar to a ``passive archive'' that is often used to compute independent metrics~\citep{pierrot2022MOQD}, but has the archive reset at every logging step (instead of persisting throughout the entire run). By doing this, we can compute a QD score and coverage value that accounts for both the quality and diversity of solutions, enabling a fair comparison between structured and unstructured approaches. To ensure that this grid does not favor the grid-based ME variants, we ensure that we project in a grid with randomized centroids.

\begin{figure*}[th]
\includegraphics[width=0.9\textwidth]{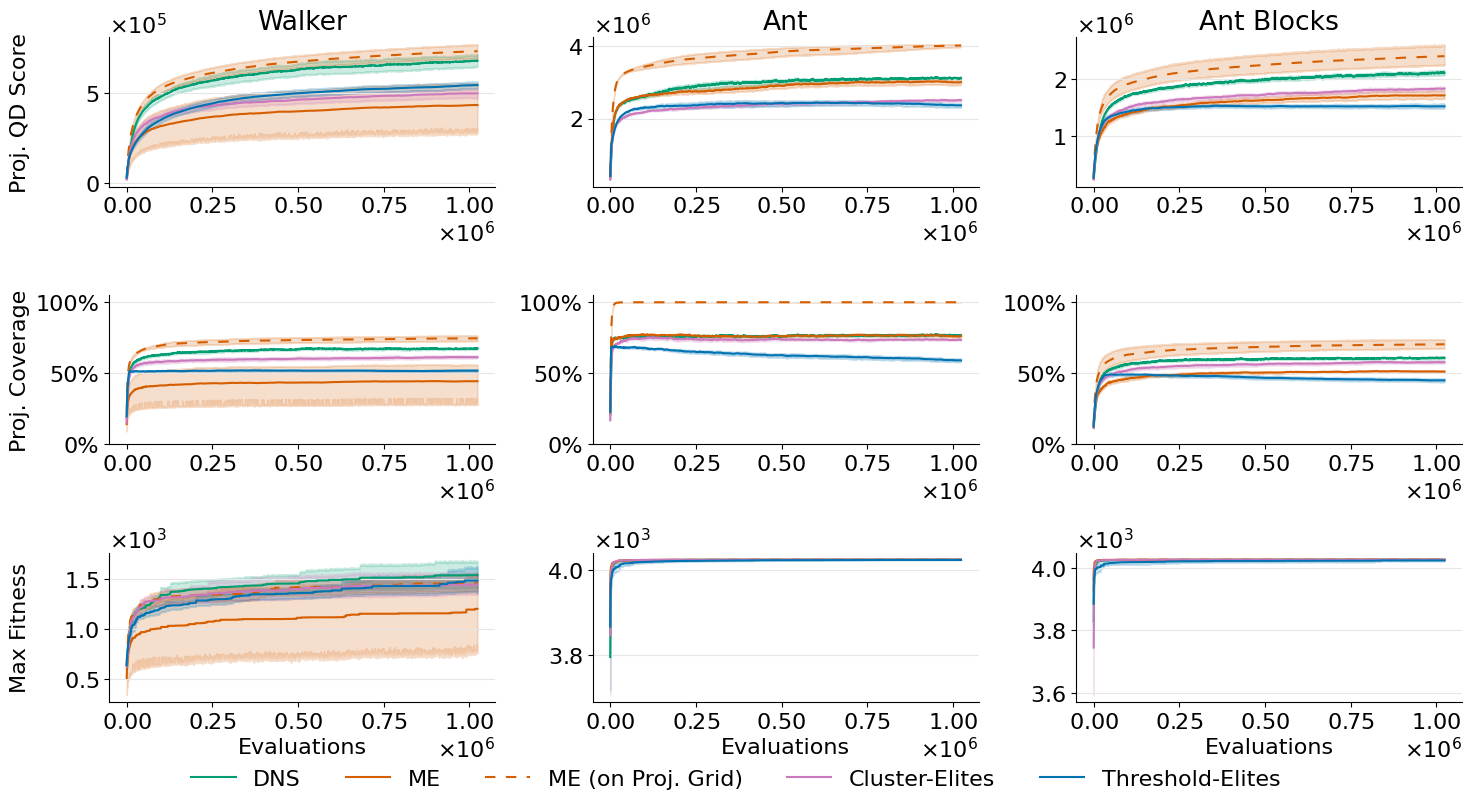}
\caption{Mujoco Environment Results. Each line represents the mean, with shaded regions representing 95\% confidence intervals. These results are aggregated across 10 independent runs. Each column represents an environment, and each row represents a given metric after projection into a random grid.}
\label{fig:results-mujoco-envs}
\end{figure*}

\subsection{Baselines}
We evaluate \oursShort{} against three established Quality-Diversity algorithms, each representing different approaches to implementing local competition in evolutionary search. We detail these baselines in the next sections.

\paragraph{MAP-Elites}
MAP-Elites~\citep{mouret_IlluminatingSearchSpaces_2015} is a foundational QD algorithm that discretizes the descriptor space into a grid, maintaining the highest-performing individual in each cell. For our experiments, we initialize a random grid for ME and project the resulting individuals onto a separate, randomly initialized grid for metric computation at each step. Additionally, we implement an optimistic variant where the same grid serves both search and evaluation, establishing an upper performance bound for grid-based approaches.

\paragraph{Cluster-Elites}
Cluster-Elites~\citep{vassiliades_ComparisonIlluminationAlgorithms_2017} is an algorithm designed to adaptively partition the descriptor space, addressing limitations of fixed-grid methods like ME. It employs a clustering mechanism to dynamically adjust centroids based on the distribution of solutions, and using these centroids as the basis to enable efficient local competition between solutions.

\paragraph{Threshold-Elites}
Threshold-Elites employs an unstructured archive (as detailed in Section~\ref{sec:background-unstructured}). This algorithm maintains diversity through a distance threshold parameter $l$, which is specifically tuned for each environment except for the unsupervised maze environment. Appendix~\ref{app:l-value} outlines how we tuned this value as well as the values we used. For the unsupervised maze environment, use container size control (CSC) \cite{csc}, which automatically determines the value for $l$. Like our approach, Threshold-Elites implements a local competition mechanism where solutions compete based on their proximity in the descriptor space, eliminating the need for explicit space partitioning. This characteristic makes Threshold-Elites particularly suitable for problems with unbounded descriptor spaces.

\subsection{Results}
In this section, we discuss the results and performance of our approach across three key conditions: 1) when the descriptor space is well defined, 2) when the descriptor space is structured but misaligned with the MAP-Elites grid, and 3) when the descriptor space is unsupervised, as in the latent space of an encoding model.

\subsubsection{\oursShort{} Outperforms Baselines in Well-Defined Descriptor Spaces}
Figure~\ref{fig:results-mujoco-envs} presents results across Mujoco continuous control environments, where descriptor bounds are well established. \oursShort{} consistently outperforms both Threshold-Elites ($p{<}10^{-9}$) and the cluster-elites population ($p{<}10^{-9})$ across all three environments in terms of QD score as well as coverage values.

Critically, \oursShort{} significantly outperforms MAP-Elites in two of the three domains: Walker and Ant Blocks ($p{<}10^{-3}$). These domains contain implicit constraints that limit reachable behaviors and therefore constrain the descriptor space. For Walker, certain foot-contact proportions (e.g.,  $(0, 0)$) are infeasible. Similarly, in Ant Blocks, obstacles prevent certain regions from being explored, making the MAP-Elites grid inefficient. \oursShort{}, in contrast, effectively adapts to these constraints, maintaining high diversity and quality.

In the Ant domain, \oursShort{} and MAP-Elites perform competitively. This can be attributed to the environment's well-defined, 2D descriptor space that directly aligns with MAP-Elites' grid. This suggests that while MAP-Elites can be effective in cases where the grid directly aligns with the descriptor space, \oursShort{} remains the superior choice when dealing with well-defined but constrained spaces.

\subsubsection{\oursShort{} is Robust to Increasing Descriptor Space Dimensionality}
\begin{figure}[th]
\includegraphics[width=1\linewidth]{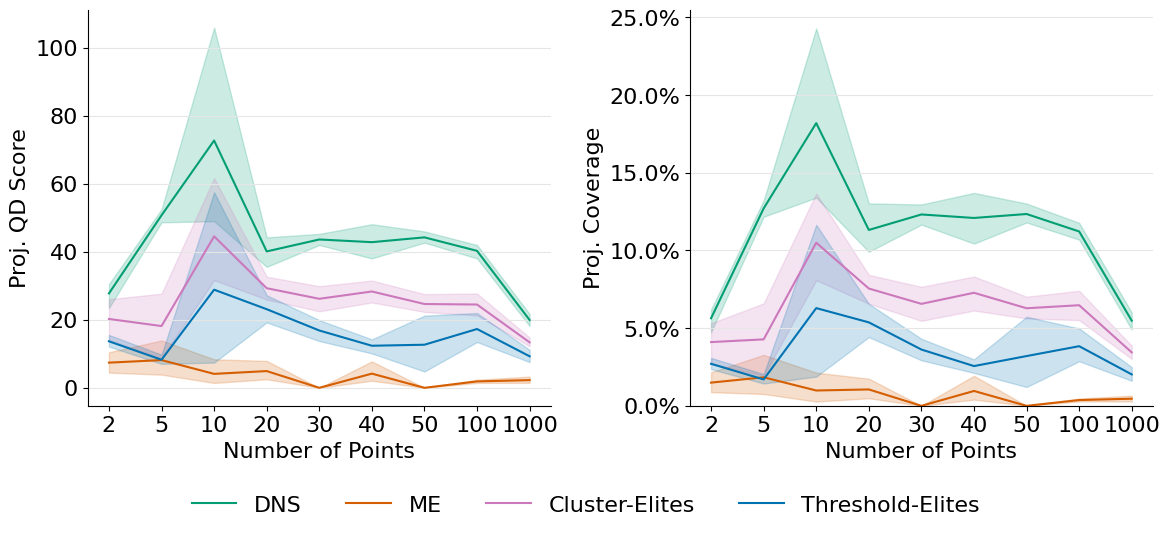}
\caption{Kheperax with Varied Descriptor Dimensionality}
\label{fig:results-kheperax-points}
\end{figure}

A key challenge in QD optimization is maintaining performance as the dimensionality of the descriptor space increases. To evaluate robustness to increasing descriptor dimensionality, we scale the descriptor space dimensionality in the Kheperax domain by sampling the robot's trajectory into $n$ evenly spaced points.

Figure~\ref{fig:results-kheperax-points} demonstrates that for all $n$ except $n{=}2$ and $n{=}10$, \oursShort{} significantly outperforms all other baselines ($p{<}0.05$) in terms of both QD Score and coverage. For $n{=}2$ and $n{=}10$, DNS outperforms Cluster-Elites and Threshold-Elites, but not significantly. DNS significantly outperforms MAP-Elites on all environment conditions $(p{<}0.05)$. Unlike MAP-Elites, which suffers from discretization inefficiencies in high-dimensional spaces, \oursShort{} effectively adapts to the growing descriptor complexity. As the dimensionality of the descriptor increases, a larger proportion of the grid represents unreachable behavior descriptors, which leads to stunting the discovery of stepping stones that are needed to explore the cells that are truly reachable. \oursShort{}, on the other hand, is able to adapt the search to the currently reachable portions of the descriptor space, leading to more effective stepping stone discovery.


\subsubsection{\oursShort{} is Effective for Unsupervised Descriptors}
\begin{figure}[th]
\includegraphics[width=1\linewidth]{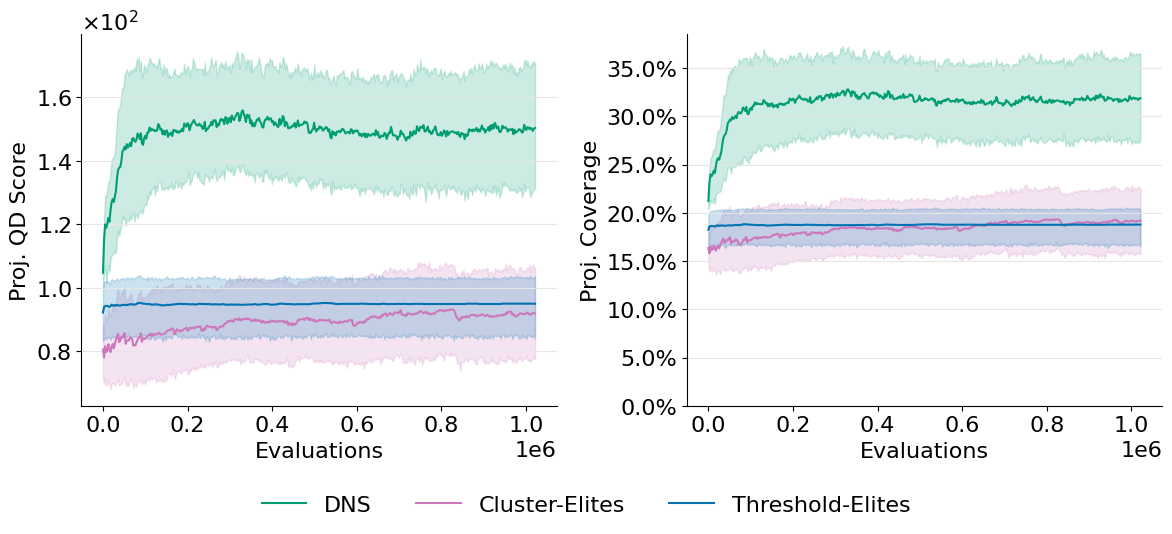}
\caption{Kheperax with Unsupervised Behavior Descriptors}
\label{fig:results-kheperax-unstructured}
\end{figure}

In environments where descriptor spaces are not predefined, \oursShort{} maintains its superior performance. Figure~\ref{fig:results-kheperax-unstructured} shows results on a variant of the maze tasks where descriptors are learned as low-dimensional embeddings of the agent's trajectory, following the AURORA algorithm~\cite{cully_AutonomousSkillDiscovery_2019}. \oursShort{} significantly outperforms other populations in terms of QD score and coverage $(p{<}0.05)$.


This highlights a fundamental advantage of \oursShort{}: it adapts to learned descriptor spaces, which are unbounded and changing throughout the evolutionary process. MAP-Elites, which requires a pre-defined and fixed grid, cannot be applied in this scenario. Furthermore, for our unsupervised Maze environment, Threshold-Elites is dependent on a complex container size control mechanism that adapts a distance parameter \cite{csc}. DNS achieves better performance with a relatively simple adaptation mechanism.

\section{Conclusion and Discussion}
In this work, we introduced \oursShort{}, a novel Quality-Diversity algorithm that fundamentally rethinks how local competition is implemented in Quality-Diversity algorithms. By eliminating the need for predefined bounds, grid structures, or fixed distance thresholds, our approach offers a more principled solution to the challenge of maintaining diversity in evolved populations.

The experimental results demonstrate that  \oursShort{} achieves state-of-the-art performance across standard QD benchmarks, validating its effectiveness as a general-purpose Quality-Diversity algorithm. Specifically,
\begin{enumerate}
    \item In well-defined descriptor spaces, \oursShort{} significantly outperforms MAP-Elites, demonstrating its ability to adapt to implicit constraints that limit reachable behaviors (e.g., Walker and Ant Blocks).
    \item In high-dimensional descriptor spaces, \oursShort{} is robust to increasing dimensionality, outperforming all baselines as complexity increases, highlighting its scalability.
    \item In unsupervised descriptor spaces, \oursShort{} significantly outperforms all methods, underscoring its ability to adapt to learned representations without requiring pre-defined bounds.
\end{enumerate}

Beyond raw performance metrics, our approach's key advantage lies in its ability to dynamically adapt to the natural structure of the solution space. This adaptability is particularly valuable in scenarios where the shape and bounds of the descriptor space cannot be known in advance, such as in unsupervised Quality-Diversity algorithms where descriptors are learned during evolution. Furthermore, \oursShort{} can serve as a drop-in replacement for grid-based mechanisms, eliminating their inherent limitations while maintaining full compatibility with both supervised and unsupervised QD approaches.

Looking forward, this work opens several promising directions for future research. First, by recasting Quality-Diversity as a genetic algorithm with local competition, we introduce a new theoretical framework where the competition function becomes a flexible, interchangeable component. This perspective allows for the development of novel local competition strategies that could further improve efficiency and effectiveness. Second, this reformulation suggests opportunities for learning optimal competition functions --- for instance, the competition mechanism could be represented as a neural network and trained to maximize population diversity and performance. As the field continues to evolve, we believe the principles introduced in this work will prove valuable in developing even more sophisticated approaches to generating diverse, high-performing solutions to complex problems.

\bibliographystyle{ACM-Reference-Format}
\bibliography{ref}

\clearpage
\appendix

\section*{Supplementary Materials}
\section{Ablation: Effect of $k$}
\begin{figure*}[th]
    \centering
    \includegraphics[width=\linewidth]{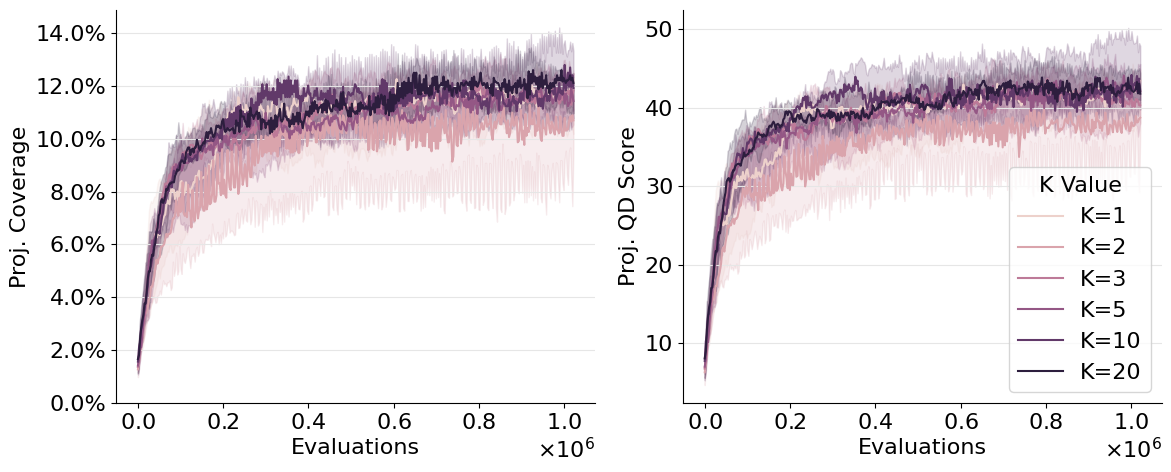}
    \caption{Results from varying $k$ for \ours{} on the Maze ($n=30$) environment. }
    \label{fig:results-kheperax-k_vary}
\end{figure*}

We run an additional experiment in order to determine the effect of the $k$ parameter on the performance of \ours{}. Figure~\ref{fig:results-kheperax-k_vary} outlines the results of this experiment. There does not appear to be a significant difference in performance for the values of $k$ studied in this work.

\section{$l$-value tuning and parameters}\label{app:l-value}

For our Threshold-Elites population runs that don't use container size control (i.e., all experiments except for the unsupervised maze environment), the value of $l$ must be carefully selected. For environments with easily characterizable behavior spaces (such as Ant and Walker based environments), the value of $l$ is simply picked to coincide with the distance between individuals in a uniformly spread grid. For the unsupervised maze, we choose an $l$-value of $0.2$. For the other maze environments, we specifically tune the $l$-value for each environmental configuration by running each maze with an $l$-value in range $\{0.0001, 0.001, 0.01, 0.1, 0.2, 0.5, 1, 2, 3\}$ and selecting the best performing configuration. The chosen $l$-values are shown in Table~\ref{tab:l_values}.

\begin{table}[h]
\caption{Chosen $l$-values for Threshold-Elites across environments.}
\label{tab:l_values}
\centering
\small
\begin{tabular}{lc}
    \toprule
    \textbf{Environment} & \textbf{Chosen $l$-value} \\
    \midrule
    Walker & $\frac{1}{2}\times\sqrt{\frac{1}{1024}}$ \\
    Ant & $\frac{1}{2}\times\sqrt{\frac{60^2}{1024}}$ \\
    Ant Blocks & $\frac{1}{2}\times\sqrt{\frac{60^2}{1024}}$ \\
    Maze ($n=2$) & 0.01 \\
    Maze ($n=5$) & 0.1 \\
    Maze ($n=10, 20, 30, 40$) & 0.5 \\
    Maze ($n=50$) & 1 \\
    Maze ($n=100$) & 1 \\
    Maze ($n=1000$) & 3 \\
    \bottomrule
\end{tabular}
\end{table}

\end{document}

%% file: notations.tex
\newcommand{\individual}{\mathbf{x}}
\newcommand{\population}{\mathbf{X}}

\newcommand{\fitness}{f}
\newcommand{\fitnesses}{\mathbf{\fitness}}

\newcommand{\descriptorSpace}{\mathcal{D}}
\newcommand{\descriptor}{\mathbf{d}}
\newcommand{\descriptors}{\descriptor}

\newcommand{\centroid}{\mathbf{c}}
\newcommand{\centroids}{\mathbf{C}}

\newcommand{\ours}{Dominated Novelty Search}
\newcommand{\oursShort}{DNS}

\newcommand{\threshold}{l}

\newcommand{\norm}[1]{\left\lVert#1\right\rVert}

\definecolor{red}{RGB}{175, 42, 20}
\definecolor{green}{RGB}{109, 158, 75}
\definecolor{blue}{RGB}{76, 115, 187}
\definecolor{purple}{RGB}{90, 64, 154}

\newtheoremstyle{numbered}
  {3pt}{3pt}{}{}{\bfseries}{.}{ }{\thmname{#1} \thmnumber{#2} \thmnote{(#3)}}

\theoremstyle{numbered}